%% file: acl_latex.tex
\pdfoutput=1

\documentclass[11pt]{article}

\usepackage[final]{acl}

\usepackage{times}
\usepackage{latexsym}

\usepackage[T1]{fontenc}

\usepackage[utf8]{inputenc}

\usepackage{microtype}

\IfFileExists{inconsolata.sty}{\usepackage{inconsolata}}{}

\usepackage{graphicx}
\usepackage{tabularx}
\usepackage{booktabs}

\input{commands}

\title{\textsc{DeliChess}: A Multi-party Dialogue Dataset for Deliberation in Chess Puzzle Solving}
\hypersetup{
  pdftitle={DeliChess: A Multi-party Dialogue Dataset for Deliberation in Chess Puzzle Solving},
  pdfauthor={Xiaochen Zhu; Georgi Karadzhov; Tom Stafford; Andreas Vlachos}
}

\newcommand{\camemailaddress}[1]{\href{mailto:#1@cam.ac.uk}{#1}}
\newcommand{\sheemailaddress}[1]{\href{mailto:#1@sheffield.ac.uk}{#1}}

\author{
  Xiaochen Zhu$^{1}$\thanks{Equal contribution.}\quad
  Georgi Karadzhov$^{1}$\footnotemark[1]\quad
  Tom Stafford$^{2}$\quad
  Andreas Vlachos$^{1}$\\
  $^{1}$University of Cambridge \quad
  $^{2}$University of Sheffield\\
  \{\camemailaddress{xz479}, \camemailaddress{gmk34},
  \camemailaddress{av308}\}@cam.ac.uk\\
  \sheemailaddress{t.stafford}@sheffield.ac.uk
}

\begin{document}
\maketitle
\begin{abstract}
Multi-party dialogue is a critical setting for studying collaborative reasoning and decision-making, yet existing datasets rarely focus on structured, reasoning-intensive tasks. We introduce \textsc{DeliChess}, a dataset of group deliberation dialogues in which participants collaboratively solve multiple-choice chess puzzles. Participants first answer independently, then engage in multi-party deliberation and revise their individual answers. The dataset comprises 107 sessions containing 321 puzzle discussions across 24 unique positions, with full transcripts, pre- and post-deliberation choices, and utterance-level annotations of communicative function, epistemic stance, and usefulness in supporting deliberation. Our analyses show that greater diversity in initial solution quality is associated with larger gains, while answer trajectories reveal how deliberation can recover, discover, or lose strong answers.
Cases in which groups surpass every independent answer are associated with sustained reasoning and epistemic openness. We further propose a diagnostic action-selection task: the tested LLMs show limited exact agreement with human-attested helpful actions, while ranked predictions reveal partial alignment. Overall, our dataset provides a testbed for modelling group reasoning, dialogue dynamics, and conditions for effective deliberation.
\end{abstract}

\input{sections/01_introduction}
\input{sections/02_related_work}
\input{sections/03_task_and_data}
\input{sections/04_annotation_schema}
\input{sections/05_analysis}
\input{sections/06_llm_action_experiment}
\input{sections/07_conclusion}

\section*{Limitations}

\textsc{DeliChess} studies short, synchronous, text-based deliberations among screened chess players choosing from five candidate moves. This controlled setting gives objective outcomes, but may not generalise to open-ended, asynchronous, face-to-face, or high-stakes deliberation. Reusing 24 positions enables comparisons across groups facing the same problems, while limiting generalisation across chess positions. Because the dialogues are textual, they also omit paralinguistic and nonverbal cues such as prosody, speech timing, gesture, and facial expression. Such cues can shape how people interpret confidence and evaluate claims \cite{guyer2021paralinguistic}, and may therefore matter for deliberation in richer communication settings. Stockfish scores measure move quality rather than every educational or interpersonal benefit of deliberation, and the candidate set does not represent unconstrained chess play.

The analyses are observational and should be read as descriptive evidence about when groups improve, not as causal intervention policies. The main corpus has one retained annotation per dialogue and cannot support corpus-wide inter-annotator reliability estimates. The training study and a 12-dialogue
reannotation audit provide quality-control evidence (Appendices~\ref{app:training-agreement} and~\ref{app:reannotation-quality}), but Deliberative Agent Usefulness remains an application-oriented judgement of the deliberative move type, not the chess correctness of the utterance.

The LLM action-selection experiment is likewise diagnostic. It evaluates agreement with one observed human move that annotators judged helpful; it does not assume that this move was the only useful intervention. Other action types might also have helped in the same context, so the model results should be interpreted as evidence of weak alignment with human-attested deliberative support rather than as a complete measure of intervention quality.

\section*{Ethics Statement}
The study was approved through the authors' institutional ethics review process. Participants were recruited through Prolific, provided informed consent, and were compensated for their time. Both compensation rates exceeded the UK National Living Wage at the time of collection; recruitment and compensation details appear in Appendices~\ref{app:data-details} and~\ref{app:annotation-collection}. The research dataset contains task responses and anonymised dialogue transcripts only; no personal identifying information is included in the corpus or released artifacts. Manual review confirms the dataset contains no hate speech, threats, or targeted abuse.

\section*{Acknowledgments}

This project was supported by a donation from Google. We thank Nemanja Vaci, Merim Bilalić and Justin Ong for informative discussion of chess and chess research.

\bibliography{custom}

\clearpage
\input{sections/appendix}

\end{document}

%% file: commands.tex
\usepackage{adjustbox}
\usepackage{multirow} 
\usepackage{listings}
\usepackage{enumitem}
\usepackage{xspace}
\usepackage{amsmath}
\usepackage{enumitem}
\usepackage{tabularx}
\usepackage{wrapfig}
\usepackage{pgfplots}
\usepackage{tcolorbox}
\tcbuselibrary{breakable,skins,listings}
\usepackage{graphicx}
\usepackage{booktabs}
\usepackage{subcaption}
\pgfplotsset{compat=1.18}

\lstdefinestyle{plainins}{
    backgroundcolor=\color{white},   
    commentstyle=\color{codegreen},
    keywordstyle=\color{magenta},
    numberstyle=\tiny\color{codegray},
    stringstyle=\color{codepurple},
    basicstyle=\ttfamily\small,
    breakatwhitespace=false,         
    breaklines=true,                 
    captionpos=b,                    
    keepspaces=true,                 
    numbers=none,                    
    numbersep=5pt,                  
    showspaces=false,                
    showstringspaces=false,
    showtabs=false,                  
    tabsize=2,
    aboveskip=0pt,
    belowskip=0pt,
    frame=single,
    escapeinside={(*@}{@*)}
}

\newcommand{\rparagraph}[1]{\vspace{1.2mm}\noindent\textbf{#1.}}

%% file: sections/01_introduction.tex
\section{Introduction}

\begin{figure*}[t]
    \centering
    \includegraphics[width=0.9\linewidth]{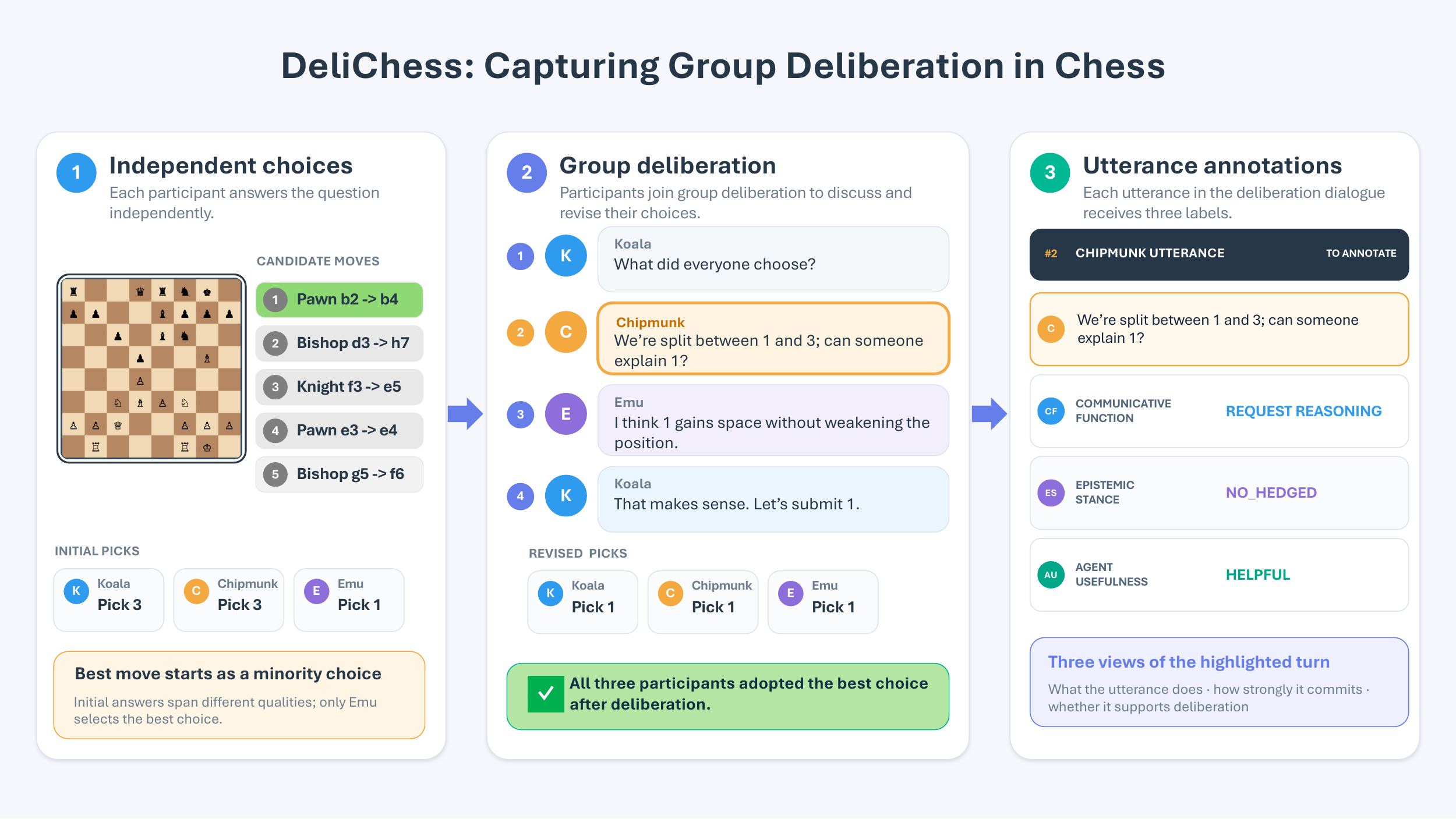}
    \caption{\textsc{DeliChess} data collection and annotation pipeline. Participants choose moves independently, deliberate, and revise their answers. Utterances are annotated for communicative function, epistemic stance, and deliberative agent usefulness.}
    \label{fig:teaser}
\end{figure*}


The \textit{``wisdom of crowds''} describes how combining diverse and
independent perspectives can produce judgements that surpass those of
individuals \cite{surowiecki2005wisdom}. Group deliberation allows individuals to exchange and integrate perspectives in pursuit of better collective decisions \cite{mercier2017enigma}. It is important across various domains including scientific peer review, political discourse, and organisational management \cite{blacksher2012public}. Research in NLP and the social sciences has therefore examined deliberative multi-party dialogues (MPDs) to understand and improve their effectiveness \cite{karpowitz2007groups,black2014methods,mercier2017natural}.

However, the benefits of deliberation are not guaranteed, as it can amplify conformity, suppress valuable minority views, and produce premature consensus \cite{janis1972victims,asch1956studies,priem1995structured}. This variability motivates conversational support that can recognise what a group needs and intervene appropriately, for example by eliciting reasons, encouraging alternative proposals, or coordinating a decision. 
Dialogue agents based on large language models (LLMs) create an opportunity to provide such support at scale, and early work suggests that LLM-mediated interaction can improve some deliberative experiences and outcomes \cite{argyle2023leveraging,tessler2024common,Behrendt2025NLP,khan2024debating}. Realising this opportunity requires models that understand not only what participants say, but also how their conversational actions contribute to the deliberative process.

However, progress is constrained by the data available for studying this behaviour. General MPD corpora, including online disputes, scripted dialogue, and casual conversations, contain rich interactions but often lack a shared decision objective, paired choices before and after deliberation, or an external measure of decision quality \cite{de2021beg,banchs2012movie,Henderson2019}. Task-oriented resources such as \textsc{StreetCrowd} \cite{niculae2016conversational} and \textsc{DeliData} \cite{karadzhov2023delidata} provide clearer outcome grounding, but use comparatively bounded laboratory tasks with limited scope for extended strategic reasoning.
Their annotation schemes also do not jointly capture a broad inventory of communicative functions and speakers' epistemic commitment. This limits our ability to study not only whether deliberation improves decisions, but also how the dynamics of deliberation, proposals, explanations, challenges, and different degrees of commitment contribute to decision outcomes. 

To address this gap, we introduce \textsc{DeliChess}, a dataset of 
deliberative multi-party dialogues over chess puzzles, with the data collection and annotation pipeline illustrated in Figure~\ref{fig:teaser}. Chess itself provides sufficient strategic depth for the effects of deliberation to become visible, while its well-defined rules and engine-based move evaluations provide an objective, graded measure of decision quality.  
In our setting, participants first solve three multiple-choice puzzles independently, then deliberate with their group and revise their individual answers. The resulting dataset contains
107 sessions containing 321 puzzle discussions
and 7,667 utterances, together with paired pre- and post-deliberation choices. We additionally introduce a multi-layer annotation scheme in which each utterance is annotated for its communicative function, epistemic stance, and usefulness as a possible action for a deliberation-support agent. These layers connect conversational processes to measurable decision outcomes while supporting the development of deliberative agents.

This design enables us to examine deliberation dynamics that cannot be captured by aggregate performance measures or previous annotation schemes. Beyond confirming that deliberation improves average group performance while producing highly variable outcomes, we show that greater diversity in initial solution quality is associated with larger gains, and trace how groups recover, discover, or lose strong answers. We find that epistemic openness, expressed through hedged stances, accompanies more productive deliberation, including 
cases in which a group surpasses every answer independently proposed. Finally, we use the annotations to construct a diagnostic action-selection task and find that the tested LLMs align only weakly with helpful actions observed in human deliberation. Overall, our study provides a foundation for studying when group reasoning succeeds and for developing conversational agents that can support it.

%% file: sections/02_related_work.tex
\section{Related Work}

\rparagraph{Deliberation Datasets and NLP}
Several multi-party dialogue corpora cover diverse interaction settings, including WikiDisputes \cite{de2021beg}, Movie-DiC \cite{banchs2012movie}, and Reddit-based datasets \cite{Henderson2019}. However, they often lack well-defined decision objectives and objective measures of outcome quality, making it difficult to assess whether deliberation improves group decisions. Task-oriented corpora provide clearer grounding. \textsc{StreetCrowd} records individual and team estimates in a collaborative geolocation game of estimating locations from street images, allowing deliberation to be evaluated against objectively scored outcomes \cite{niculae2016conversational}. \textsc{DeliData} captures MPD alongside pre- and post-deliberation performance measures and a probing-oriented annotation scheme grounded in the Wason card-selection task, a controlled logic problem in which participants identify the cards needed to test a conditional rule \cite{wason1968reasoning,karadzhov2023delidata}. This framework has been extended to the group evaluation of AI-generated text \cite{lee2025collaborativeevaluationdeepfaketext}, while \citet{nath2024any} introduce deliberation chains that model the causal structure underlying probing questions in collaborative dialogue. Together, these studies demonstrate the broader analytical value of deliberation-grounded datasets. Nevertheless, the underlying decision problems in collected MPDs remain relatively constrained, involving short reasoning chains and few alternatives. Existing datasets therefore provide limited coverage of the strategic planning, branching possibilities, uncertainty management, and multi-step inference central to complex group decision-making.

\rparagraph{Multi-Party Dialogue Benchmarks}
Existing benchmarks examine complementary aspects of multi-party interaction: conversation disentanglement and reply structure \cite{kummerfeld-etal-2019-large}, discourse-grounded reading comprehension in Molweni \cite{li-etal-2020-molweni}, and addressee recognition and response selection \cite{penzo-etal-2024-llms}. \textsc{DeliChess} connects conversational behaviour to individual decisions before and after deliberation and graded measures of decision quality. Its action-selection task examines which facilitative move follows a discussion context, complementing these structural and comprehension tasks and work on information-seeking policies in task-oriented dialogue \cite{geishauser2021informationgain}.

\rparagraph{Annotating Deliberative Dialogue}
The probing-oriented annotation scheme used in the \textsc{DeliData} line of work \cite{karadzhov2023delidata,nath2024any} focuses on whether utterances probe for information. While useful for studying information elicitation, it does not provide a general communicative-function inventory that distinguishes, for example, proposals, explanations, challenges, evaluations, and decision-coordination moves. Such distinctions become important in studying the details of deliberation dynamics, where contributions can play different roles in developing and evaluating competing solutions. General dialogue-act frameworks such as DAMSL \cite{allen1997damsl} and ISO 24617-2 \cite{bunt2020iso} provide foundations for distinguishing communicative functions, including informing, requesting, responding, and managing interaction. Communicative function alone, however, does not capture how speakers position themselves toward a claim. Stance-taking involves evaluation, positioning, and alignment \cite{dubois2007stance}, while speakers' expressed commitment and relative epistemic authority shape how their claims function in interaction \cite{heritage2012epistemics}. Linguistic markers of epistemic stance can also vary across tasks and speakers \cite{gablasova2017epistemic}. Together, these strands motivate a multidimensional account of deliberative dialogue that separates what an utterance does from how strongly its speaker commits to it. Existing deliberation datasets do not jointly capture these dimensions or explicitly represent conversational moves as potential interventions for deliberation-support agents.

\rparagraph{Chess as a Reasoning Domain}
Chess is an established domain in cognitive science for studying expertise, problem-solving, and decision-making under constraints \cite{bilalic2008inflexibility,bilalic2008good,bilalic2010mechanism}. Previous work has primarily examined individual reasoning through psychometric, behavioural, and response-time analyses \cite{sunde2022speed,van2005psychometric}. In contrast, collaborative reasoning through chess dialogue remains comparatively under-explored. Compared with existing MPD corpus settings, chess requires multi-step planning across branching alternatives, managing uncertainty about consequences, and reconciling substantively different strategic evaluations, making it suitable for analysing reasoning-heavy deliberation.


%% file: sections/03_task_and_data.tex
\section{Task and Data Collection}

\rparagraph{Task design}
We use chess puzzles as a structured testbed for group deliberation. We compile a pool of 24 unique puzzles from the Amsterdam Chess Test \cite{van2005psychometric}, evenly divided among endgame, positional, and tactical problems. Endgame puzzles require precise play in reduced-material positions; positional puzzles reward gradual improvement without an immediate tactical gain; and tactical puzzles require forcing sequences that win material or deliver checkmate. These categories allow us to examine deliberation across different reasoning demands. For each puzzle, we identify five candidate moves: the best move identified by Stockfish \cite{Stockfish2025} and four plausible distractors. Distractor selection is informed by human move frequencies and Stockfish evaluations, producing alternatives that span a range of move qualities without being obviously incorrect. Each dialogue session is randomly assigned one puzzle from each category, and for each puzzle participants select among the five candidate moves.

\rparagraph{Participants and procedure}
Following the pre-deliberation/discussion/post-deliberation design of \textsc{DeliData} \cite{karadzhov2023delidata}, participants were recruited through Prolific\footnote{\url{https://www.prolific.com/}} and screened for basic chess competence using a paid qualification task. In the study, participants first solved all three puzzles independently. They were then assigned to synchronous group deliberation, where they could examine the board position, compare candidate moves, and revise their choices. After deliberation, each participant submitted their final individual answers. This design allows us to compare individual decisions before and after deliberation while also observing the conversational process through which groups negotiate their answers.

\rparagraph{Corpus construction}
The final corpus contains 107 multi-party dialogue sessions, each covering three puzzles, for a total of 321 puzzle discussions across 24 unique positions. We use \emph{dialogue} for the complete session and \emph{dialogue--puzzle case} for one puzzle discussion. The dataset is available online.\footnote{\url{https://huggingface.co/datasets/SpaceHunterInf/DeliChess}} Groups contain two to five active speakers, with an average of 3.41 participants and 71.65 utterances per dialogue. The event logs record both initial solo choices and revised post-deliberation choices.

\begin{table}[t]
\centering
\small
\setlength{\tabcolsep}{4pt}
\begin{tabular}{p{0.67\linewidth}r}
\toprule
\textbf{Corpus statistic} & \textbf{Value} \\
\midrule
Dialogues & 107 \\
Dialogue--puzzle cases & 321 \\
Annotated utterances & 7,667 \\
Mean speakers per dialogue & 3.41 $\pm$ 0.81 \\
Mean utterances per dialogue & 71.65 $\pm$ 34.76 \\
Mean utterances per puzzle discussion & 23.88 \\
Matched pre- and post-deliberation choices & 1,057 \\
Mean dialogue duration (min) & 18.80 $\pm$ 9.12 \\
\bottomrule
\end{tabular}
\caption{Corpus statistics. A choice is included in the matched set only when the same participant--puzzle observation is available both before and after deliberation.}
\label{tab:dataset-overview}
\end{table}

\rparagraph{Outcome measures}
We evaluate move quality using three bounded measures derived from Stockfish evaluations, each scoring the chosen move among the five candidates. \textbf{Simple} assigns a linearly decreasing reward based on the engine ranking of the chosen move. \textbf{Reciprocal Rank (RR)}, defined as $1/r$ for engine rank $r$, gives substantially greater weight to selecting the top-ranked move. \textbf{WDL} is independent of the ordinal ranking and instead uses Stockfish's win--draw--loss probability estimates to compute the expected game-outcome score for the chosen move, capturing absolute positional quality. 

Choices remain individual throughout the task. Group performance for a puzzle is the mean score across its matched participants, and overall dialogue performance is the mean across the dialogue's three puzzles. Participant-level performance gain is defined as the post-deliberation score minus the matched pre-deliberation score; group-level gain averages these individual changes within each puzzle case and, for overall analyses, across the three puzzles. 
Additional details on engine configuration, puzzle-level statistics, and sensitivity analyses with raw engine scores are provided in Appendix~\ref{app:data-details}--\ref{app:stockfish-evaluation}.

%% file: sections/04_annotation_schema.tex
\section{Annotation Schema and Collection}
\label{sec:annotation}

To move beyond outcome-only analysis and support investigation of deliberation dynamics, we annotate each chat utterance along three complementary dimensions (see Figure~\ref{fig:teaser}): \textbf{Communicative Function (CF)}, \textbf{Epistemic Stance (ES)}, and \textbf{Deliberative Agent Usefulness (AU)}. CF captures what an utterance does in the deliberation, ES captures how the speaker displays commitment to a task-relevant stance, and AU captures whether the same kind of conversational move could be useful for an AI agent facilitating deliberation. 
Importantly, annotators were not shown the chess board and were instructed not to assess whether participants' chess judgements were correct; all labels concern the conversational role of the utterance, not its chess accuracy. Table~\ref{tab:annotation-inventory} summarises the label inventory; full definitions, decision rules, and worked examples are provided in Appendix~\ref{app:annotation-guidelines}.

\begin{table}[t]
\centering
\small
\setlength{\tabcolsep}{3.5pt}
\renewcommand{\arraystretch}{1.13}
\begin{tabular}{@{}p{0.12\columnwidth}p{0.76\columnwidth}r@{}}
\toprule
\textbf{Dim.} & \textbf{Labels} & \textbf{$|L|$} \\
\midrule
CF &
Answer proposal; reasoning; alternative exploration; reasoning request; evaluation/critique; agreement/alignment; decision coordination; participation management; social/off-task
& 9 \\
\midrule
ES &
No task stance; hedged task stance; unhedged task stance
& 3 \\
\midrule
AU &
Helpful; neutral; not applicable; harmful
& 4 \\
\bottomrule
\end{tabular}
\caption{Annotation label inventory. CF = Communicative Function; ES = Epistemic Stance; AU = Deliberative Agent Usefulness. $|L|$ denotes the number of labels.}
\label{tab:annotation-inventory}
\end{table}

\rparagraph{Communicative Function}
CF identifies the dominant conversational action performed by each utterance. Where an utterance serves multiple functions, for instance, proposing an answer while simultaneously providing reasoning, annotators select the primary function based on the utterance's main contribution to the deliberation. This primary-function representation supports consistent annotation; ES separately records whether the move is hedged. Secondary communicative functions remain a useful target for future multilabel annotation. The labels distinguish answer proposals, reasoning, exploration of alternatives, requests for justification, critique or evaluation, agreement, decision coordination, participation management, and social or off-task interaction. This dimension is motivated by dialogue-act frameworks, which treat utterances as actions performed in interaction \cite{allen1997damsl,bunt2020iso}.

\rparagraph{Epistemic Stance}
ES records whether an utterance expresses a task-relevant stance and, if so, whether that stance is linguistically hedged. We distinguish utterances with no task stance, hedged task stances, and unhedged task stances. This dimension captures displayed commitment rather than chess expertise or correctness. The classification is based on the presence or absence of hedging markers: expressions containing markers such as ``maybe'', ``I think'', or question framing are treated as hedged, while any task stance without such markers, including bare answer proposals such as ``4'', is treated as unhedged. The distinction draws on work on stance-taking and epistemics in interaction \cite{dubois2007stance,heritage2012epistemics,gablasova2017epistemic}, but uses a deliberately simple label set to support reliable annotation of chat utterances.

\rparagraph{Deliberative Agent Usefulness}
AU provides an application-oriented judgement of whether the same kind of conversational action could be useful as an intervention by a dialogue agent, complementing the descriptive CF and ES layers. A move is labelled helpful if it could support reasoning, exploration, participation, coordination, or consensus-building; neutral if it is task-related but adds little deliberative value; not applicable if it falls outside the agent's role, such as pure greetings or personal answer choices; and harmful if it would likely bias the group, shut down disagreement, or push premature consensus. This dimension is motivated by prior work on deliberative dialogue agents and conversational interventions \cite{karadzhov2023delidata,nath2024any,argyle2023leveraging}, but differs from task-performance annotation: annotators judge the deliberative function of the action, not its chess correctness or exact wording. For example, an utterance such as ``I don't like this move, knight is in a weak position, maybe let's try another one'' contains personal judgement that might not be correct in game, but the underlying action, prompting the group to explore alternatives, is helpful.

\rparagraph{Annotation procedure and quality control}
Annotators were recruited through Prolific and paid for training using written guidelines and calibration dialogues. Training included practice annotations on held-out dialogues followed by review; only annotators who demonstrated sufficient agreement with reference labels were retained for the main annotation task. 
The final corpus contains annotations for 7,667 genuine utterances, each assigned to one retained annotator after training and quality-control filtering. To monitor annotation quality, we inserted one synthetic attention-check utterance into each dialogue and analysed these separately from genuine chat; all 107 attention checks in the final corpus matched the expected labels on all three dimensions. We additionally conducted a 12-dialogue reannotation audit, obtaining Cohen's $\kappa$ scores of 0.715 for CF, 0.923 for ES, and 0.610 for AU, with corresponding macro-F1 scores of 0.717, 0.934, and 0.721. Further details on recruitment, training, selective retention, compensation, workload, and reannotation quality control are provided in Appendix~\ref{app:annotation-collection}--\ref{app:reannotation-quality}. Condensed annotation guidelines appear in Appendix~\ref{app:annotation-guidelines}; the complete instructions shown to annotators are included in the supplementary material.

%% file: sections/05_analysis.tex
\section{Results}
\label{sec:results}

We analyse 107 dialogue sessions containing 321 puzzle discussions to ask whether group performance changes after deliberation, which starting conditions are associated with larger gains, and how groups move toward better or worse answers. 
Performance gains are tested against zero with two-sided paired $t$-tests on the corresponding dialogue-level before/after values. Unless otherwise stated, we report 95\% confidence intervals and use Benjamini--Hochberg false discovery rate (FDR) correction within each analysis family; \(q\) denotes the resulting adjusted \(p\)-value. 
Fine-grained process analyses are correlational and should not be interpreted as causal intervention effects.

\subsection{Deliberation Improves Performance, but Unevenly}

\input{sections/generated/performance_gain_table}

Group performance is higher after deliberation on all three bounded measures (Table~\ref{tab:performance-gains}; all overall $q<0.001$). Mean group RR performance rises from 0.473 before deliberation to 0.552 afterwards, a gain of $\Delta=0.079$, with comparable gains under Simple and WDL scores. Gains are largest and most consistent for tactical puzzles. Positional puzzles also show positive gains, while endgame gains are weaker and mainly visible under WDL.

Aggregate improvement after group deliberation has already been documented in other deliberation tasks, including Wason selection \cite{karadzhov2023delidata}. Prior work on collaborative deepfake-text evaluation likewise emphasizes that deliberation produces highly variable outcomes across groups \cite{lee2025collaborativeevaluationdeepfaketext}. \textsc{DeliChess} exhibits both patterns: performance improves on average, but RR increases in 157 of the 321 dialogue--puzzle cases, remains unchanged in 50, and decreases in 114. This heterogeneity is not merely noise around an average effect. It motivates closer study of when deliberation helps, when it harms, and which starting conditions and conversational processes distinguish those outcomes. We therefore use the richer task record to examine what information groups begin with and how their answers change.

\subsection{Solution-Quality Diversity Predicts Larger Gains}

\begin{figure*}[t]
\centering
\includegraphics[width=0.78\textwidth]{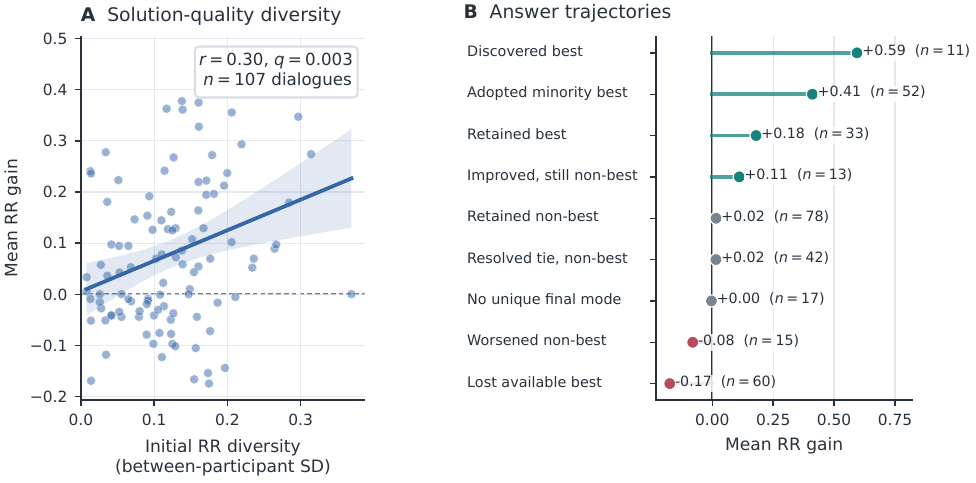}
\caption{\textbf{Initial diversity and answer trajectories.} (A) Groups with greater dispersion in initial RR performance show larger gains; shading gives the 95\% confidence band. (B) Exhaustive classification of the 321 dialogue--puzzle cases by how the final modal answer relates to the initial answers. Point labels report mean RR gain and number of cases.}
\label{fig:diversity-decisions}
\vspace{-3mm}
\end{figure*}

We first ask whether group performance gain is associated with initially diverse information. We define initial solution-quality diversity by averaging each participant's solo score across the three puzzles and then taking the between-participant standard deviation. Greater initial diversity in solution quality is associated with larger group performance gains: this holds for RR diversity ($r=0.295$, $q=0.003$; Figure~\ref{fig:diversity-decisions}A) and WDL diversity ($r=0.273$, $q=0.007$), while the weaker Simple-score association does not survive correction. By contrast, diversity in the \emph{identity} of initial moves is unrelated to RR gain ($r=0.076$, $p=0.437$). Thus, the relevant variation is in the quality of participants' initial solutions rather than disagreement alone. These correlations are descriptive and do not establish that diversity itself causes improvement.

\subsection{Answer Trajectories Reveal Knowledge Recovery and Loss}

The paired pre- and post-deliberation choices allow us to distinguish how groups arrive at their final answers rather than reducing each deliberation to a single gain score. Figure~\ref{fig:diversity-decisions}B classifies all 321 dialogue--puzzle cases by the relationship between the initial answer pool and the derived final modal answer. In 52 cases, the final mode is the Stockfish-best candidate initially held only by a minority, producing a large mean RR gain ($+0.411$). In another 11 cases, the unique final mode is a Stockfish-best candidate that nobody initially selected ($+0.594$). These 11 cases form a subset of the 19 strict-assembly cases discussed in Section~\ref{sec:strict-assembly}. Deliberation can therefore surface underused knowledge and sometimes produce a solution absent from the initial pool.

The same trajectories expose information loss. In 60 cases, at least one participant initially selected a Stockfish-best candidate, but the final mode moved away from it ($-0.174$). Although these losses are slightly more frequent than minority-best recoveries (60 vs. 52 cases), their mean RR magnitude is less than half as large (0.174 vs. 0.411). The most common outcome is retaining a non-best answer. Nor is increased agreement sufficient by itself: among cases where the final modal share increases, 143 show positive RR gain while 102 show negative gain. These trajectories explain why deliberation is high variance: interaction can recover minority knowledge, construct a new answer, or suppress the strongest answer already available.

Across the three annotation dimensions, counts of providing reasoning, requesting reasoning, exploring alternatives, hedged stances and helpful deliberation utterances are positively associated with RR gain (all \(q\leq0.012\); Appendix~\ref{app:annotation-associations}). However, none of their proportions shows a reliable independent association after adjusting for dialogue length and baseline RR. Improving deliberations therefore contain more task-relevant conversational activity, but the present analysis does not identify any single move as an independent driver of improvement. We consequently examine hedging in the more specific context of strict assembly.

\subsection{Epistemic Openness and Strict Assembly}
\label{sec:strict-assembly}

The Epistemic Stance layer lets us examine whether tentative commitment is part of constructive deliberation. At the dialogue level, the number of hedged utterances is positively associated with RR gain ($r=0.278$, $q=0.012$). However, the association does not remain reliable when hedging is represented as a proportion and the model adjusts for dialogue length and baseline performance ($\beta=0.027$, $q=0.791$; Appendix~\ref{app:annotation-associations}). Hedging should therefore not be interpreted as a general intervention that independently improves performance. Its clearest association appears in a more specific constructive outcome.

Adapting the assembly-bonus concept of \citet{collins1964social}, we define \emph{strict assembly} as a puzzle case where the derived final modal answer has a better Stockfish rank than every answer initially supplied by the participants. Unlike ordinary improvement, strict assembly cannot be explained by adopting the strongest initial answer: the group has moved beyond its initial solution set. Strict assembly occurs in 19 of the 162 eligible dialogue--puzzle cases where no participant initially chose the Stockfish-best candidate. 11 of these 19 cases end with the best candidate as the unique final mode, as described above; the other 8 end with a non-best candidate that nevertheless outranks every initially supplied answer.

\begin{figure*}[t]
\centering
\includegraphics[width=0.88\textwidth]{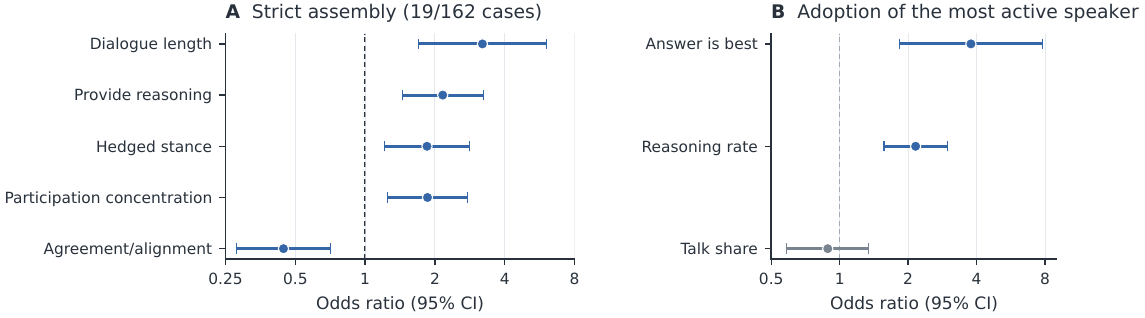}
\caption{Exploratory adjusted models of (A) strict assembly and (B) adoption of the most active speaker's answer. Intervals use dialogue-clustered uncertainty.}
\label{fig:useful-mechanisms}
\end{figure*}

Across separate logistic models adjusting for baseline puzzle RR, the quality of the strongest initial answer, group size, and puzzle type, strict assembly is more likely in cases with longer deliberation, more reasoning and hedging, more concentrated participation, and less agreement/alignment (Figure~\ref{fig:useful-mechanisms}A). In particular, the association with hedging remains reliable after multiple-comparison correction (odds ratio [OR] $=1.85$, $q=0.009$). These rare cases are characterised by sustained reasoning, tentative claims, and delayed alignment, a pattern we describe as \emph{epistemic openness}. Given only 19 positive cases, we treat these results as exploratory evidence about possible mechanisms rather than as general prescriptions for facilitation.

Panel B addresses whether participation concentration merely reflects uncalibrated dominance. Among participants who initially disagree with the most active speaker, switching toward that speaker is more likely when the speaker's answer is best and when the speaker provides more reasoning, while raw talk share is inconclusive. Productive influence therefore appears more closely calibrated to answer and reasoning quality than to talk share alone, although the evidence remains observational. Additional uptake, prediction, and robustness details appear in Appendix~\ref{app:uptake-definition}--\ref{app:influence-models}.

%% file: sections/generated/performance_gain_table.tex
\begin{table}[t]
\centering
\scriptsize
\setlength{\tabcolsep}{3pt}
\resizebox{\columnwidth}{!}{%
\begin{tabular}{@{}l r r r@{}}
\toprule
\textbf{Scope} & \textbf{Simple $\Delta$ [95\% CI]} & \textbf{RR $\Delta$ [95\% CI]} & \textbf{WDL $\Delta$ [95\% CI]} \\
\midrule
Overall & \textbf{+0.065 [+0.041, +0.089]} & \textbf{+0.079 [+0.051, +0.108]} & \textbf{+0.086 [+0.058, +0.114]} \\
Endgame & +0.031 [-0.002, +0.064] & +0.023 [-0.017, +0.062] & \textbf{+0.065 [+0.021, +0.109]} \\
Positional & \textbf{+0.074 [+0.037, +0.111]} & \textbf{+0.098 [+0.053, +0.143]} & \textbf{+0.062 [+0.030, +0.094]} \\
Tactical & \textbf{+0.091 [+0.047, +0.134]} & \textbf{+0.118 [+0.072, +0.163]} & \textbf{+0.130 [+0.075, +0.184]} \\
\bottomrule
\end{tabular}
}
\caption{Mean group performance gain after deliberation. Overall estimates average each dialogue's three puzzles before testing ($n=107$); type-specific estimates use one puzzle per dialogue. Bold entries remain significant after Benjamini--Hochberg correction.}
\label{tab:performance-gains}
\end{table}

%% file: sections/06_llm_action_experiment.tex
\section{LLM Action-Selection Experiment}
\label{sec:llm-action-selection}

The preceding analyses suggest that productive deliberation involves more than contributing a correct answer: conversational actions also structure how groups reason over alternatives and coordinate decisions. We therefore ask whether current LLMs can identify the contextually attested helpful action in an ongoing deliberation.

\rparagraph{Task} Each instance consists of a dialogue prefix followed by a human utterance annotated as helpful under Deliberative Agent Usefulness. We represent the target action using the utterance's Communicative Function (CF) label, drawn from the inventory defined in Section~\ref{sec:annotation} and summarized in Table~\ref{tab:annotation-inventory}. Given only the dialogue prefix, the model must return one of its six facilitative action types: providing reasoning, exploring alternatives, requesting reasoning, evaluating or critiquing an option, coordinating a decision, or managing participation. We exclude direct answer proposals, bare agreement, and off-task interaction because the task concerns support for the deliberative process rather than participation as another voter. The target is therefore a context-attested helpful action: it records one action that a human produced and annotators judged helpful, without assuming that it was the only appropriate intervention.

\rparagraph{Data}
We evaluate the models on a balanced diagnostic sample of 60 dialogue-prefix instances drawn from 52 dialogues, with 10 instances for each of the six facilitative action labels. We sample these instances from 97 eligible dialogues containing at least one puzzle that improved after deliberation and at least one helpful facilitative utterance. Appendix~\ref{app:action-selection-sampling} provides the precise eligibility and sampling criteria. Because the sample is deliberately filtered and label-balanced, the results diagnose action-selection agreement rather than estimate model performance under the corpus's naturally occurring label distribution.

\rparagraph{Models and evaluation} We compare a majority CF baseline with Gemini 3.5 Flash \citep{google2026geminiModels}, Gemma 4 31B IT \citep{gemmateam2026gemma4technicalreport}, and Tencent Hunyuan Hy3 \citep{tencentHunyuan2026Github}. In the few-shot setting, the prompt contains six demonstrations, one for each action label, drawn from dialogues excluded from the evaluation sample. We report strict accuracy, Cohen's $\kappa$, and macro-F1. Invalid or missing outputs remain in the denominator and are counted as incorrect. The zero- and six-shot prompts are reproduced in Appendix~\ref{app:action-prompts}.

\begin{table}[!ht]
\centering
\small
\setlength{\tabcolsep}{4pt}
\begin{tabularx}{\columnwidth}{@{}Xrrr@{}}
\toprule
\textbf{Model} & \textbf{Acc.} & \textbf{$\kappa$} & \textbf{Macro-F1} \\
\midrule
\multicolumn{4}{@{}l}{\emph{Baseline}} \\
Majority baseline & 0.167 & 0.000 & 0.048 \\
\addlinespace
\multicolumn{4}{@{}l}{\emph{Zero-shot LLMs}} \\
Gemini 3.5 Flash & 0.350 & 0.220 & 0.319 \\
Gemma 4 31B IT & 0.283 & 0.151 & 0.278 \\
Tencent Hy3 & 0.250 & 0.103 & 0.165 \\
\addlinespace
\multicolumn{4}{@{}l}{\emph{Few-shot LLMs ($k=6$)}} \\
Gemini 3.5 Flash & 0.267 & 0.120 & 0.223 \\
Gemma 4 31B IT & 0.300 & 0.160 & 0.275 \\
Tencent Hy3 & 0.333 & 0.203 & 0.305 \\
\bottomrule
\end{tabularx}
\caption{Strict performance on context-attested helpful-action prediction ($n=60$). Invalid or missing outputs are counted as incorrect.}
\label{tab:llm-action-selection}
\end{table}

\rparagraph{Results}
Table~\ref{tab:llm-action-selection} shows low exact agreement with the communicative functions of the observed helpful utterances. Gemini 3.5 Flash performs best in the zero-shot setting, reaching 0.350 accuracy, $\kappa=0.220$, and macro-F1 of 0.319, but no model exceeds 0.350 accuracy in either setting. Few-shot prompting with one balanced demonstration per label does not consistently improve performance: Gemini declines to 0.267 accuracy, while Gemma and Hunyuan improve only modestly.

\rparagraph{Ranked action selection}
We also asked each model to rank all six actions for the same 60 contexts. Across models and prompting conditions, the attested action appears in the top two positions in 40--55\% of cases and the top three in 60--75\%, compared with random-ranking baselines of 33.3\% and 50\%. Mean reciprocal rank ranges from 0.460 to 0.570 (random: 0.408). Ranking reveals partial alignment beyond exact matching, with no consistent benefit from few-shot prompting. Appendix~\ref{app:action-ranking} provides results and configuration details. Both evaluations measure alignment with human-attested actions; assessing the acceptability of other interventions requires separate human judgements.

%% file: sections/07_conclusion.tex
\section{Conclusion}

We introduced \textsc{DeliChess}, a dataset linking 107 multi-party sessions comprising 321 puzzle discussions with paired individual decisions, engine-based move quality, and utterance-level annotations. Deliberation improves performance on average, but outcomes vary substantially across groups. Greater diversity in initial solution quality is associated with larger gains, while answer trajectories reveal the recovery, discovery, and loss of strong solutions. Rare strict-assembly cases, in which groups surpass every initially available solution, co-occur with sustained reasoning and greater epistemic openness. In a diagnostic action-selection task, the tested LLMs show limited exact agreement with human-attested helpful actions, while ranking reveals partial alignment across the candidate actions. We hope that \textsc{DeliChess} will support future work on modelling how interaction transforms distributed knowledge and on developing agents that facilitate, rather than merely participate in, deliberation.

%% file: sections/appendix.tex
\appendix
\section{Supplementary Material}
\label{sec:appendix}

\subsection{Comparison with Existing Task-Oriented Corpora}
\label{app:corpus-comparison}

Table~\ref{tab:corpus-comparison} situates \textsc{DeliChess} alongside
StreetCrowd \citep{niculae2016conversational}, the Settlers of Catan (SoC)
corpus \citep{afantenos2012settlers}, and \textsc{DeliData}
\citep{karadzhov2023delidata}. The earlier corpus counts follow the published
comparison in \citet{karadzhov2023delidata}. \textsc{DeliChess} combines objective outcome measurement with individual
choices both before and after deliberation and three complementary
utterance-level annotation layers. In particular, its epistemic-stance and
agent-usefulness annotations capture information not represented in the
comparison corpora.

\begin{table}[!ht]
\centering
\footnotesize
\setlength{\tabcolsep}{1.5pt}
\renewcommand{\arraystretch}{1.08}
\begin{tabularx}{\columnwidth}{@{}Xcccc@{}}
\toprule
\textbf{Property} & \shortstack{\textbf{Street}\\\textbf{Crowd}} & \textbf{SoC} & \shortstack{\textsc{\textbf{Deli}}\\\textsc{\textbf{Data}}} & \shortstack{\textsc{\textbf{Deli}}\\\textsc{\textbf{Chess}}} \\
\midrule
Dialogues & 1,450 & 32 & 500 & 107 \\
Utterances & 17,545 & 2,512 & 14,003 & 7,667 \\
Utterances/dialogue & 12.1 & 78.5 & 28.0 & 71.7$^{*}$ \\
Collaborative decision & Yes & No & Yes & Yes \\
Objective outcome & Distance & Game & Task & Engine \\
Post-delib. individual choice & No & No & Yes & Yes \\
Dialogue acts / CF & No & \shortstack{Speech\\acts} & 5 roles & 9 CFs \\
Epistemic stance & No & No & No & 3 labels \\
Agent usefulness & No & No & No & 4 labels \\
\bottomrule
\end{tabularx}
\caption{Comparison with task-oriented multi-party dialogue corpora.
$^{*}$Each \textsc{DeliChess} dialogue covers three puzzles: 71.7 utterances per
session corresponds to 23.9 per puzzle discussion, compared with 28.0 per
single-task dialogue in \textsc{DeliData}. ``Game'' denotes competitive game
outcome rather than a collaborative decision-quality measure. \textsc{DeliData}'s five role labels form part of its
probing-oriented deliberation annotation.}
\label{tab:corpus-comparison}
\end{table}

\subsection{Dialogue Participant Screening and Collection}
\label{app:data-details}

Table~\ref{tab:puzzle-type-stats} reports means and sample standard deviations across all 107 sessions. Chat events have no puzzle label, so we infer the puzzle type from the next move submitted during group play. Five utterances have no subsequent move submission and are counted as unassigned. All 7,667 utterances are included in Table~\ref{tab:dataset-overview}. Time is the elapsed task time assigned to each puzzle stage.

Dialogue participants were recruited through Prolific. Eligibility required
participants to be adults, English-speaking, and to have a Prolific approval
rate of at least 98\%. The paid qualification task contained three chess
puzzles; participants qualified when the mean Stockfish rank of their selected
moves was no worse than third. Qualified participants were invited to the
dialogue study, which paid \pounds5 and took 18.8 minutes on average (an
effective rate of \pounds15.96 per hour). Across
the 107 retained dialogues, the logs contain 365 distinct active-speaker
identifiers, with two to five active speakers per dialogue; no identifier
occurs in more than one retained dialogue.

\begin{table}[!ht]
\centering
\small
\setlength{\tabcolsep}{4pt}
\begin{tabular}{@{}lrr@{}}
\toprule
\textbf{Puzzle} & \textbf{Messages} & \textbf{Time (min)} \\
\midrule
Endgame    & 23.12 $\pm$ 14.16 & 6.23 $\pm$ 5.82 \\
Positional & 25.80 $\pm$ 17.63 & 7.23 $\pm$ 5.91 \\
Tactical   & 22.68 $\pm$ 12.53 & 5.34 $\pm$ 5.84 \\
Unassigned & 0.05 $\pm$ 0.21 & -- \\
\bottomrule
\end{tabular}
\caption{Mean message count and elapsed time by puzzle type.}
\label{tab:puzzle-type-stats}
\end{table}

\subsection{Engine Evaluation and Performance Sensitivity}
\label{app:stockfish-evaluation}
\label{app:additional-performance}

We rescore candidate positions using Stockfish 17.1 from the original mover's
perspective, with one thread, 128 MB hash, and a fixed search budget of 100,000
nodes per position. Candidate ranks are assigned from these fresh evaluations.
WDL expectation is computed as $p_W + 0.5p_D$ from Stockfish's win--draw--loss
output. Raw centipawn evaluations are converted to pawns; mate scores are used
for ranking but are retained only in the raw-evaluation sensitivity analysis.

Across the 107 dialogues, raw engine evaluation has a group performance gain
of $-0.944$ pawns (95\% CI $[-14.832,12.945]$, $q=0.893$). This unstable
estimate reflects extreme tactical evaluations and motivates our use of
bounded outcomes. At the dialogue--puzzle level, raw evaluation increases in
174 of 321 cases, remains unchanged in 50, and decreases in 97. These counts
are reported only as a sensitivity breakdown; the primary Results section uses
RR throughout.

\subsection{Annotation Recruitment and Collection}
\label{app:annotation-collection}

Fourteen Prolific candidates completed a paid one-hour training session using
the written guidelines and a shared 34-utterance calibration dialogue. We
ranked their initial labels against the reference using exact three-label
agreement, with mean initial Cohen's $\kappa$ as the tie-breaker, and selected
and hired the highest-ranked 12 for corpus annotation. Each paid annotation
session contained two dialogues, took 57 minutes on average, and was
compensated at \pounds13.45 per hour. Annotators could complete multiple
sessions. Eleven of the 12 selected annotators contributed to the final
107-dialogue corpus; the median contribution was eight dialogues per annotator
(range 2--20). The 11 final annotators were proficient English speakers
residing across ten countries on five continents. Residence information was
self-reported through Prolific, was not used for selection or analysis, and
will not be released with the corpus. 

Each dialogue received one final annotation. One synthetic attention-check
utterance was inserted into each annotation task and was analysed separately
from genuine dialogue content. All 107 attention checks in the final corpus
matched the expected CF, ES, and AU labels; the synthetic rows are excluded
from the 7,667-utterance linguistic analyses.

\subsection{Annotation Training and Revision}
\label{app:training-agreement}

For the 12 selected candidates described above, training agreement is computed
over the same 34 genuine utterances. For cross-candidate comparison, we join
every response to the current reference by message ID.

\input{sections/generated/training_agreement_table}

Mean agreement across dimensions rises from 65.4\% initially to 92.3\% after
reference-guided revision, while mean $\kappa$ rises from 0.494 to 0.884. Exact
agreement on all three labels rises from 35.3\% to 83.8\%. The initial scores
measure independent agreement with the reference; the final scores measure
responsiveness after candidates viewed it and must not be interpreted as
independent inter-annotator reliability. Early candidates saw a snapshotted
reference, but recomputing the aggregates against the version each candidate
actually saw changes every agreement estimate by less than one percentage
point.

\subsection{Reannotation Quality Audit}
\label{app:reannotation-quality}

As an additional quality-control check, we collected an independent
reannotation for 12 of the 107 final corpus dialogues and compared the
reannotation labels against the retained corpus labels on genuine utterances.
Table~\ref{tab:reannotation-quality} reports Cohen's $\kappa$ and macro-F1 for
each annotation dimension, using the retained corpus labels as the reference
for the F1 calculation.

\begin{table}[t]
\centering
\footnotesize
\setlength{\tabcolsep}{3pt}
\begin{tabularx}{\columnwidth}{@{}Xrr@{}}
\toprule
\textbf{Dimension} & \textbf{Cohen's $\kappa$} & \textbf{Macro-F1} \\
\midrule
Communicative Function (CF) & 0.715 & 0.717 \\
Epistemic Stance (ES)      & 0.923 & 0.934 \\
Deliberative Agent Usefulness (AU) & 0.610 & 0.721 \\
\bottomrule
\end{tabularx}
\caption{Agreement in a 12-dialogue reannotation audit. Scores compare the
independent reannotation with the retained corpus labels for genuine
utterances only.}
\label{tab:reannotation-quality}
\end{table}

The audit provides targeted evidence that the annotation scheme was applied
consistently beyond the initial calibration dialogue. Agreement is strongest
for ES and substantial for CF. AU shows lower chance-corrected agreement, which
is expected for a more application-oriented and partly subjective judgement,
but the macro-F1 score indicates that annotators still recover the main AU
distinctions reasonably well. We therefore treat the annotations as reliable
descriptive process measures while avoiding corpus-wide inter-annotator
reliability claims, since the remaining dialogues retain one final annotation.

\subsection{Annotation Label Distributions}
\label{app:annotation-distributions}

\begin{figure*}[t]
    \centering
    \includegraphics[width=0.94\textwidth]{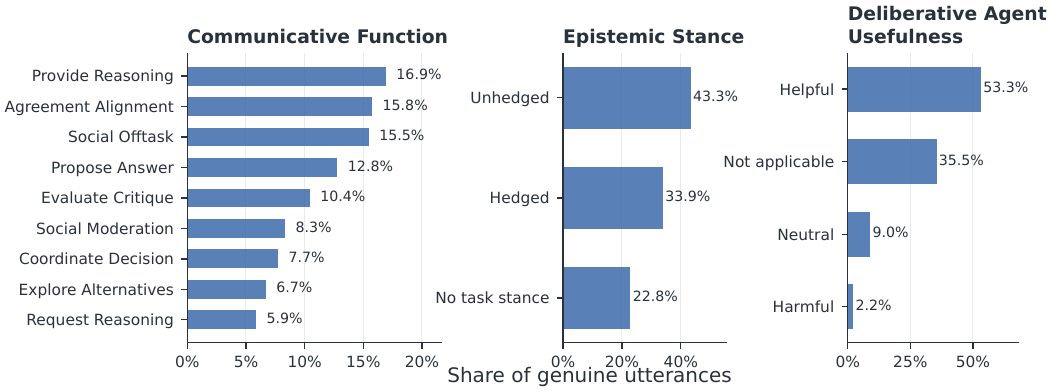}
    \caption{Human-label distributions over 7,667 genuine utterances.}
    \label{fig:annotation-distributions}
\end{figure*}

Figure~\ref{fig:annotation-distributions} reports the complete prevalence of
every label over genuine chat utterances.

\subsection{Condensed Annotation Guidelines}
\label{app:annotation-guidelines}

Annotators answered three questions for each utterance in its local dialogue
context:
\begin{enumerate}
    \item \textbf{Communicative Function:} What is the utterance's primary
    conversational action?
    \item \textbf{Epistemic Stance:} Does it express a task-relevant stance,
    and is that stance hedged or uncertain?
    \item \textbf{Deliberative Agent Usefulness:} Would this kind of move be
    useful for an AI agent supporting the group's deliberation?
\end{enumerate}

Detailed annotation labels and definitions are provided in
Table~\ref{tab:annotation-definitions}.

\begin{table*}[t]
\centering
\scriptsize
\renewcommand{\arraystretch}{1.08}
\begin{tabularx}{\textwidth}{@{}p{2.5cm}p{4.0cm}X@{}}
\toprule
\textbf{Dimension} & \textbf{Label} & \textbf{Operational definition} \\
\midrule
CF & \texttt{PROPOSE\_ANSWER} & Suggests a candidate move or answer. \\
 & \texttt{PROVIDE\_REASONING} & Explains why a move or option may work or fail. \\
 & \texttt{EXPLORE\_ALTERNATIVES} & Introduces, compares, or keeps open multiple options. \\
 & \texttt{REQUEST\_REASONING} & Requests explanation, justification, or clarification. \\
 & \texttt{EVALUATE\_CRITIQUE} & Assesses, supports, rejects, or compares candidate answers. \\
 & \texttt{AGREEMENT\_ALIGNMENT} & Signals agreement, acceptance, or alignment. \\
 & \texttt{COORDINATE\_DECISION} & Manages submission, task progress, or the final decision. \\
 & \texttt{SOCIAL\_MODERATION} & Manages participation, inclusion, or turn-taking. \\
 & \texttt{SOCIAL\_OFFTASK} & Performs greeting, closing, humour, or unrelated social talk. \\
\midrule
ES & \texttt{NO\_TASK\_STANCE} & Expresses no task-relevant stance in context. \\
 & \texttt{HEDGED} & Expresses a task stance marked as uncertain or tentative. \\
 & \texttt{NO\_HEDGED} & Expresses a task stance without uncertainty marking. \\
\midrule
AU & \texttt{HELPFUL} & Could help an agent support reasoning, exploration, participation, or coordination. \\
 & \texttt{NEUTRAL} & Task-related but adds little new deliberative value. \\
 & \texttt{NOT\_APPLICABLE} & Outside the deliberation agent's role, including personal choices and social talk. \\
 & \texttt{HARMFUL} & Could bias, derail, exclude, or prematurely close deliberation. \\
\bottomrule
\end{tabularx}
\caption{Operational definitions for all annotation labels.}
\label{tab:annotation-definitions}
\end{table*}

\begin{table*}[t]
\centering
\footnotesize
\begin{tabular}{@{}p{4.7cm}p{3.8cm}p{2.8cm}p{3.4cm}@{}}
\toprule
\textbf{Utterance} & \textbf{Communicative Function} &
\textbf{Epistemic Stance} & \textbf{Deliberative Agent Usefulness} \\
\midrule
``Hi everyone'' & \texttt{SOCIAL\_OFFTASK} &
\texttt{NO\_TASK\_STANCE} & \texttt{NOT\_APPLICABLE} \\
``What did everyone choose?'' & \texttt{SOCIAL\_MODERATION} &
\texttt{NO\_HEDGED} & \texttt{HELPFUL} \\
``I think 4 works because the king gets trapped'' &
\texttt{PROVIDE\_REASONING} & \texttt{HEDGED} & \texttt{HELPFUL} \\
``4'' & \texttt{PROPOSE\_ANSWER} & \texttt{NO\_HEDGED} &
\texttt{NOT\_APPLICABLE} \\
``We seem split between 2 and 4; can someone explain 4?'' &
\texttt{REQUEST\_REASONING} & \texttt{HEDGED} & \texttt{HELPFUL} \\
``Let's submit 4'' & \texttt{COORDINATE\_DECISION} &
\texttt{NO\_HEDGED} & \texttt{HELPFUL} if appropriately timed \\
``Let's just submit 4; no need to discuss'' &
\texttt{COORDINATE\_DECISION} & \texttt{NO\_HEDGED} & \texttt{HARMFUL} \\
\bottomrule
\end{tabular}
\caption{Representative examples from the annotation guidelines.}
\label{tab:annotation-examples}
\end{table*}

\subsection{Conversational-Uptake Definitions}
\label{app:uptake-definition}

We examine whether particular conversational moves receive locally relevant
uptake from other speakers. The uptake analysis is computed over genuine
utterances in event order; synthetic attention checks are excluded. For every
trigger, we inspect the next three genuine utterances and count the trigger as
receiving uptake if at least one qualifying response is produced by a different
speaker. A trigger can contribute at most one success, using the first
qualifying response in its window. Windows may overlap, and the same response
can provide uptake for more than one preceding trigger. The reported rate pools
all successful triggers and all triggers across the 107 dialogues rather than
averaging dialogue-level rates.

The five trigger--response mappings are:
\begin{enumerate}[leftmargin=*,itemsep=0pt,topsep=2pt]
    \item social moderation followed by a task response or decision
    coordination;
    \item exploration followed by evaluation, critique, or reasoning;
    \item a request for reasoning followed by reasoning;
    \item reasoning followed by agreement or alignment; and
    \item reasoning or critique followed by decision coordination.
\end{enumerate}
These are descriptive temporal co-occurrences: the calculation does not
establish that the trigger elicited the response or that the transition
affected performance.

\begin{figure}[t]
\centering
\includegraphics[width=\columnwidth]{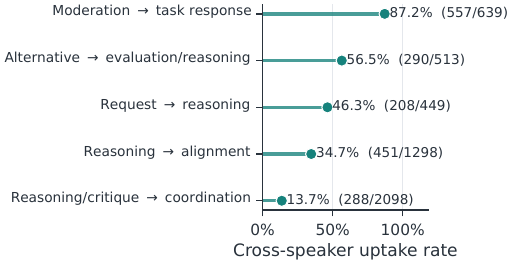}
\caption{Cross-speaker uptake within the next three utterances. Counts are
observed uptakes/triggers.}
\label{fig:conversational-uptake}
\end{figure}

\subsection{Annotation Associations}
\label{app:annotation-associations}

\begin{figure*}[t]
\centering
\includegraphics[width=0.92\textwidth]{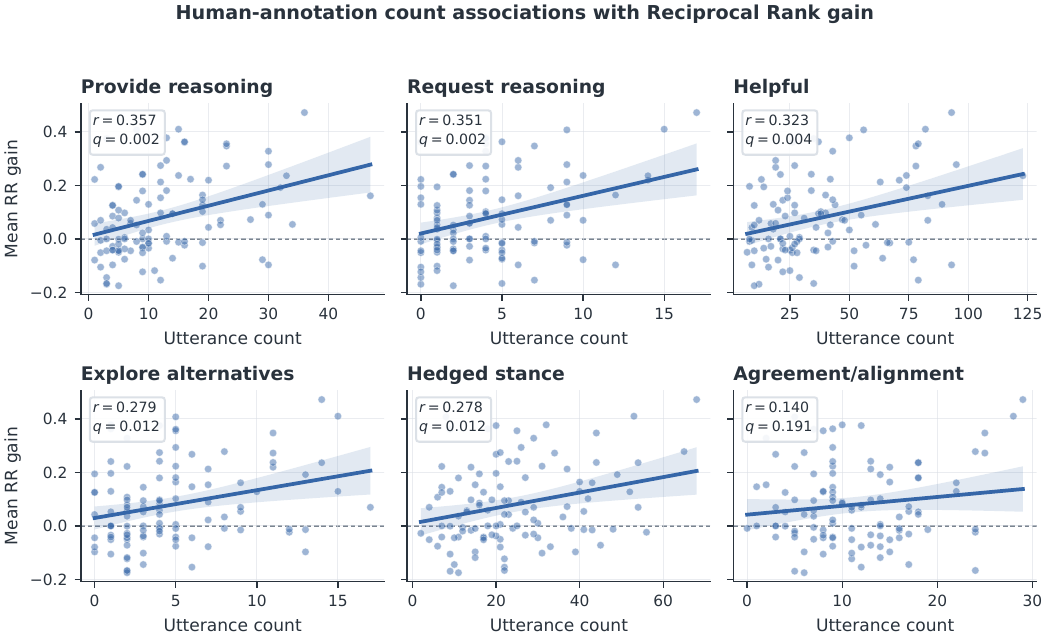}
\caption{Strongest raw human-annotation count associations with dialogue-level
RR gain. Each point is one dialogue; lines show simple linear fits with
confidence bands.}
\label{fig:top-annotation-counts-rr-gain}
\end{figure*}

Figure~\ref{fig:top-annotation-counts-rr-gain} visualises the strongest raw
count associations with RR gain. Table~\ref{tab:annotation-associations}
contrasts these raw count associations with models of annotation proportions
adjusted for log dialogue length and baseline RR. Counts of several
task-relevant behaviours covary with improvement, but none has an independently
reliable adjusted association. This is why the main text does not interpret any
single label as a general intervention effect.

\begin{table}[!ht]
\centering
\scriptsize
\setlength{\tabcolsep}{3pt}
\begin{tabularx}{\columnwidth}{@{}Xrrr@{}}
\toprule
\textbf{Feature} & \textbf{Count $r$ ($q$)} & \textbf{Adjusted $\beta$ [95\% CI]} & \textbf{$q$} \\
\midrule
Provide reasoning    & 0.357 (0.002) &  0.165 [$-0.058$, 0.388] & 0.613 \\
Request reasoning    & 0.351 (0.002) &  0.143 [$-0.072$, 0.359] & 0.613 \\
Helpful              & 0.323 (0.004) &  0.103 [$-0.091$, 0.297] & 0.722 \\
Explore alternatives & 0.279 (0.012) &  0.083 [$-0.113$, 0.279] & 0.722 \\
Hedged stance        & 0.278 (0.012) &  0.027 [$-0.154$, 0.208] & 0.791 \\
Agreement/alignment  & 0.140 (0.191) & $-0.164$ [$-0.348$, 0.021] & 0.613 \\
\bottomrule
\end{tabularx}
\caption{Associations with dialogue-level RR gain ($n=107$). Adjusted
coefficients are standardised and use HC3 standard errors. For both the count
correlations and adjusted-proportion models, $q$ corrects across all 16 CF, ES,
and AU labels within the RR family.}
\label{tab:annotation-associations}
\end{table}

\subsection{General Outcome Discrimination}
\label{app:outcome-discrimination}

We use 200 repetitions of five-fold held-out evaluation to distinguish 63
improved from 40 worsened dialogues; four unchanged dialogues are excluded.
All imputation and scaling are fitted within the training folds. The full
annotation and process feature sets do not improve over the compact activity
model (Table~\ref{tab:dialogue-prediction}).

\begin{table}[!ht]
\centering
\scriptsize
\begin{tabularx}{\columnwidth}{@{}Xrr@{}}
\toprule
\textbf{Features} & \textbf{ROC AUC} & \textbf{Average precision} \\
\midrule
Starting conditions             & 0.581 $\pm$ 0.026 & 0.711 \\
$+$ dialogue activity           & 0.612 $\pm$ 0.023 & 0.764 \\
$+$ annotation proportions      & 0.591 $\pm$ 0.033 & 0.706 \\
$+$ annotations and process     & 0.584 $\pm$ 0.035 & 0.688 \\
\bottomrule
\end{tabularx}
\caption{Repeated held-out discrimination of improved versus worsened
dialogues. Uncertainty is the standard deviation across repetitions.}
\label{tab:dialogue-prediction}
\end{table}

\subsection{Strict Assembly Models}
\label{app:strict-assembly}

Strict assembly models are restricted to 162 eligible dialogue--puzzle cases across 91 dialogues for
which no initial answer is Stockfish-best; 19 produce a strict bonus. Each row
in Table~\ref{tab:strict-assembly-models} is a separate logistic model adjusted
for puzzle baseline RR, the best initial RR, group size, and puzzle type.
Continuous features are standardised, standard errors are clustered by
dialogue, and $q$ corrects across the full 12-feature family.

\begin{table}[!ht]
\centering
\scriptsize
\setlength{\tabcolsep}{4pt}
\begin{tabularx}{\columnwidth}{@{}Xrr@{}}
\toprule
\textbf{Feature} & \textbf{OR [95\% CI]} & \textbf{$q$} \\
\midrule
Dialogue length             & 3.21 [1.69, 6.06] & 0.002 \\
Provide reasoning           & 2.16 [1.45, 3.23] & 0.002 \\
Agreement/alignment         & 0.44 [0.28, 0.71] & 0.003 \\
Participation concentration & 1.86 [1.24, 2.78] & 0.007 \\
Hedged stance               & 1.85 [1.22, 2.81] & 0.009 \\
Helpful                     & 1.42 [0.97, 2.07] & 0.147  \\
Request reasoning           & 1.38 [0.92, 2.06] & 0.210  \\
Early exploration           & 1.31 [0.81, 2.14] & 0.408  \\
Evaluate/critique           & 1.13 [0.77, 1.66] & 0.654  \\
Harmful                     & 0.86 [0.53, 1.40] & 0.654  \\
Coordinate decision         & 1.15 [0.58, 2.30] & 0.748  \\
Explore alternatives        & 0.96 [0.55, 1.70] & 0.898  \\
\bottomrule
\end{tabularx}
\caption{Adjusted exploratory models of strict assembly. ORs are per
standard-deviation increase.}
\label{tab:strict-assembly-models}
\end{table}

Repeated held-out evaluation uses 200 repetitions of dialogue-grouped
five-fold cross-validation, with all preprocessing fitted within training
folds. Table~\ref{tab:strict-assembly-prediction} shows that the compact
targeted profile generalises better than puzzle starting conditions, while an
expanded feature set does not.

\begin{table}[!ht]
\centering
\scriptsize
\begin{tabularx}{\columnwidth}{@{}Xrr@{}}
\toprule
\textbf{Features} & \textbf{ROC AUC} & \textbf{Avg. precision} \\
\midrule
Puzzle starting conditions & 0.553 $\pm$ 0.035 & 0.163 \\
$+$ dialogue activity & 0.733 $\pm$ 0.029 & 0.328 \\
$+$ targeted dialogue profile & 0.765 $\pm$ 0.024 & 0.331 \\
$+$ expanded dialogue features & 0.689 $\pm$ 0.031 & 0.290 \\
\bottomrule
\end{tabularx}
\caption{Repeated held-out discrimination of strict assembly among 162
eligible best-absent dialogue--puzzle cases (19 positives).}
\label{tab:strict-assembly-prediction}
\end{table}

\subsection{Calibrated Influence Models}
\label{app:influence-models}

The most-active-speaker analysis contains 487 switching opportunities across
100 dialogues in which another participant initially disagrees with that
speaker. The joint model adjusts for whether the speaker initially holds the
modal answer, initial modal share, baseline RR, group size, and puzzle type,
with dialogue-clustered standard errors. In this full sample
(Table~\ref{tab:influence-models}; Figure~\ref{fig:useful-mechanisms}B),
switching is more likely when the most active speaker initially holds a best
answer (OR $=3.78$) and when that speaker has a higher reasoning rate
(OR $=2.16$), whereas talk share is inconclusive (OR $=0.89$).

The same qualitative pattern holds in a robustness analysis restricted to 171
opportunities across 55 dialogues in which the most active speaker initially
opposes a unique modal answer: holding a best answer (OR $=5.64$) and reasoning
rate (OR $=3.53$) remain positively associated with switching, whereas talk
share remains inconclusive (OR $=1.77$). In a separate modal-coalition model,
reasoning share predicts switching (OR $=2.07$, $q=0.003$), while raw talk
share and hedged or unhedged stance do not.

\begin{table}[!ht]
\centering
\scriptsize
\setlength{\tabcolsep}{4pt}
\begin{tabularx}{\columnwidth}{@{}Xrr@{}}
\toprule
\textbf{Predictor} & \textbf{OR [95\% CI]} & \textbf{$p$} \\
\midrule
Most-active speaker's answer is best & 3.78 [1.84, 7.76] & $<0.001$ \\
Most-active speaker's reasoning rate & 2.16 [1.57, 2.97] & $<0.001$ \\
Most-active speaker's talk share     & 0.89 [0.59, 1.34] & 0.570 \\
\bottomrule
\end{tabularx}
\caption{Full-sample joint model of switching toward the most active speaker. Continuous
predictors are standardised.}
\label{tab:influence-models}
\end{table}

\subsection{Gemini--Human Annotation Agreement}
\label{app:gemini-human-agreement}

As a diagnostic of whether current LLMs could substitute for the human
annotation pipeline, we prompted Gemini 3.5 Flash
\citep{google2026geminiModels} with the canonical annotation guidelines and
dialogue context to label all 7,667 genuine utterances. We used temperature
zero and compared its predictions with the retained human corpus labels.
Table~\ref{tab:gemini-human-agreement} reports observed agreement, Cohen's
$\kappa$, and macro-F1; confidence intervals resample whole dialogues to
respect the nested data structure.

\begin{table}[!ht]
\centering
\scriptsize
\setlength{\tabcolsep}{3pt}
\begin{tabularx}{\columnwidth}{@{}Xrrr@{}}
\toprule
\textbf{Dimension} & \textbf{Agreement} & \textbf{$\kappa$ [95\% CI]} & \textbf{Macro-F1} \\
\midrule
Communicative Function & 0.576 & 0.514 [0.485, 0.544] & 0.540 \\
Epistemic Stance & 0.700 & 0.522 [0.477, 0.567] & 0.686 \\
Deliberative Agent Usefulness & 0.655 & 0.409 [0.370, 0.448] & 0.415 \\
All three labels (exact) & 0.355 & -- & -- \\
\bottomrule
\end{tabularx}
\caption{Gemini--human agreement over 7,667 utterances in 107 dialogues.
Intervals use 1,000 dialogue-cluster bootstrap samples.}
\label{tab:gemini-human-agreement}
\end{table}

Observed agreement is moderate, but the chance-corrected and macro-averaged
scores reveal substantial remaining differences, especially for Deliberative
Agent Usefulness. Exact agreement on the full three-label annotation occurs
for only 35.5\% of utterances. Gemini also recovers the rare \texttt{HARMFUL}
label poorly (0.6\% recall). Because the retained human labels are not an
utterance-level gold standard, these figures should be read as model--human
agreement rather than model accuracy. Nevertheless, they do not support
replacing the trained human annotators: model labels may be useful for
sensitivity analysis or annotation assistance, but still require human
oversight for corpus construction.

\input{sections/action_ranking}

\input{sections/action_selection_prompts}

%% file: sections/generated/training_agreement_table.tex
\begin{table}[!ht]
\centering
\scriptsize
\setlength{\tabcolsep}{1.8pt}
\begin{tabularx}{\columnwidth}{@{}Xrrrr@{}}
\toprule
& \multicolumn{2}{c}{\textbf{Agreement}} & \multicolumn{2}{c}{\textbf{Cohen's $\kappa$}} \\
\cmidrule(lr){2-3}\cmidrule(l){4-5}
\textbf{Dimension} & \textbf{Initial} & \textbf{Final} & \textbf{Initial} & \textbf{Final} \\
\midrule
Communicative Function & 62.3\% & 92.9\% & 0.554 & 0.916 \\
Epistemic Stance & 73.8\% & 93.6\% & 0.570 & 0.894 \\
Deliberative Agent Usefulness & 60.3\% & 90.4\% & 0.357 & 0.842 \\
Mean across dimensions & 65.4\% & 92.3\% & 0.494 & 0.884 \\
All three labels exact & 35.3\% & 83.8\% & -- & -- \\
\bottomrule
\end{tabularx}
\caption{Agreement with the current reference for the 12 selected candidates ($34$ utterances each) before and after training.}
\label{tab:training-agreement}
\end{table}

%% file: sections/action_ranking.tex
\subsection{Ranked Action-Selection Sensitivity Analysis}
\label{app:action-ranking}

We reuse the 60 evaluation contexts from 52 dialogues, with ten targets per
facilitative CF label, and ask each model to rank all six actions. Hit@$k$
measures whether the human-attested action appears among the first $k$
positions; mean reciprocal rank (MRR) averages the reciprocal of its rank.
The few-shot condition uses the same six single-action demonstrations as the
original experiment. All 360 final responses contain each action exactly once.
Table~\ref{tab:action-ranking} reports point estimates. The supplementary
ranking results include 95\% percentile confidence intervals from 10,000
bootstrap samples resampling whole dialogues (seed 23). For a uniform random
ranking, expected Hit@$k = k/6$ and expected MRR is
$\frac{1}{6}\sum_{r=1}^{6}1/r = 0.408$.

Generation uses temperature zero and seed 17. Gemini 3.5 Flash uses a thinking
budget of zero; Gemma 4 31B IT omits that unsupported option. Tencent Hy3 uses
reasoning effort \texttt{none} and an output budget of 1,024 or 2,048 tokens.
The single-action runs used \texttt{tencent/hy3:free}. That route was unavailable for ranking, so the ranking
experiment uses \texttt{tencent/hy3} through OpenRouter. These settings are
reported separately because ranking is a new generation task rather than a
rescoring of the original single-action predictions.

\begin{table}[t]
\centering
\small
\setlength{\tabcolsep}{3pt}
\begin{tabularx}{\columnwidth}{@{}Xrrrr@{}}
\toprule
\textbf{Model} & \textbf{Hit@1} & \textbf{Hit@2} & \textbf{Hit@3} & \textbf{MRR} \\
\midrule
Random ranking & 0.167 & 0.333 & 0.500 & 0.408 \\
\addlinespace
\multicolumn{5}{@{}l}{\emph{Zero-shot LLMs}} \\
Gemini 3.5 Flash & 0.300 & 0.550 & 0.750 & 0.547 \\
Gemma 4 31B IT & 0.333 & 0.500 & 0.650 & 0.546 \\
Tencent Hy3 & 0.233 & 0.450 & 0.650 & 0.483 \\
\addlinespace
\multicolumn{5}{@{}l}{\emph{Few-shot LLMs ($k=6$)}} \\
Gemini 3.5 Flash & 0.300 & 0.500 & 0.650 & 0.527 \\
Gemma 4 31B IT & 0.367 & 0.500 & 0.733 & 0.570 \\
Tencent Hy3 & 0.217 & 0.400 & 0.600 & 0.460 \\
\bottomrule
\end{tabularx}
\caption{Ranked action-selection performance ($n=60$).
Confidence intervals are provided in the supplementary ranking results.}
\label{tab:action-ranking}
\end{table}

The attested action appears in the top three for 60--75\% of contexts, showing
alignment beyond first-choice agreement. Few-shot demonstrations improve some
Gemma measures but do not consistently help across models. These rankings
measure the position of one human-attested helpful action; establishing sets
of acceptable interventions would require judgements of the alternatives.

%% file: sections/action_selection_prompts.tex
\raggedbottom
\subsection{Action-Selection Sampling and Prompts}
\label{app:action-selection-sampling}
\label{app:action-prompts}

The full action-selection pool contains 3,313 utterances annotated as helpful
whose CF label is one of the six facilitative actions defined in
Section~\ref{sec:llm-action-selection}. We restrict sampling to 97 dialogues
that contain at least one such utterance and at least one puzzle with a
positive matched pre/post change in raw engine evaluation. This raw-score sign
is used only to focus the diagnostic sample; it is not a primary performance
outcome. The balanced evaluation sample contains 60 instances from 52
dialogues, with 10 instances per action label. The six-shot prompts use one
demonstration per label from six additional dialogues excluded from the
evaluation sample. Invalid or missing model outputs remain in the denominator
and count as incorrect.

The same textual prompts were used for all tested LLMs. Each target contains
the ten preceding genuine chat turns, or all available turns when fewer than
ten precede it. The six-shot condition adds one fixed demonstration per action
label; all demonstration dialogues are excluded from evaluation. API-specific
structured-output constraints enforce the same six-label inventory and
single-key JSON response.

The prompt components are reproduced below. The zero-shot prompt concatenates
the shared instruction block and the zero-shot target block. The six-shot
prompt concatenates the shared block, the demonstration preamble and six
demonstrations, and the six-shot target block. Angle brackets mark the only
item-specific substitution.

\newtcblisting{promptblock}[2][]{
  enhanced,
  breakable,
  listing only,
  colback=blue!2,
  colframe=blue!45!black,
  colbacktitle=blue!45!black,
  coltitle=white,
  fonttitle=\bfseries,
  title={#2},
  arc=1.5mm,
  boxrule=0.5pt,
  left=2mm,
  right=2mm,
  top=1.5mm,
  bottom=1.5mm,
  before skip=7pt,
  after skip=9pt,
  listing options={
    basicstyle=\ttfamily\fontsize{8pt}{9.2pt}\selectfont,
    columns=fullflexible,
    breaklines=true,
    breakatwhitespace=false,
    keepspaces=true,
    showstringspaces=false
  },
  #1
}

\newtcblisting{demoblock}[2][]{
  enhanced,
  listing only,
  colback=black!1,
  colframe=black!35,
  colbacktitle=black!8,
  coltitle=black,
  fonttitle=\bfseries,
  title={#2},
  arc=1.5mm,
  boxrule=0.45pt,
  left=2mm,
  right=2mm,
  top=1.2mm,
  bottom=1.2mm,
  before skip=6pt,
  after skip=7pt,
  listing options={
    basicstyle=\ttfamily\fontsize{7.7pt}{8.8pt}\selectfont,
    columns=fullflexible,
    breaklines=true,
    breakatwhitespace=false,
    keepspaces=true,
    showstringspaces=false
  },
  #1
}

\begin{promptblock}{Shared instruction block}
You are evaluating a multi-party chess-puzzle deliberation.

Task: choose the Communicative Function label for the next move a deliberation-support agent should make.
Use the existing DeliChess Communicative Function definitions. Choose the action type, not a chess answer.
The agent should support reasoning, exploration, inclusion, coordination, or consensus-building without acting as another voter.

Allowed action labels:
- PROVIDE_REASONING: Explains why a move or option may work or fail.
- EXPLORE_ALTERNATIVES: Introduces, compares, or keeps open multiple candidate options.
- REQUEST_REASONING: Requests explanation, justification, or clarification.
- EVALUATE_CRITIQUE: Assesses, supports, rejects, or compares candidate answers.
- COORDINATE_DECISION: Manages submission, task progress, or the final decision.
- SOCIAL_MODERATION: Manages participation, inclusion, or turn-taking.
\end{promptblock}

\begin{promptblock}[colback=green!2,colframe=green!40!black,colbacktitle=green!40!black]{Zero-shot target block}
Conversation so far:
<TARGET DIALOGUE PREFIX>

Output exactly one compact JSON object. Do not include markdown, prose, or explanations.
Return valid JSON only:
{"action_label": "<one allowed label>"}
\end{promptblock}

\rparagraph{Six-shot demonstrations ($k=6$)}

The following block is inserted after the allowed-label definitions. Each
card's title is supplied here for readability; the actual prompt uses the
numbered \texttt{Example} and \texttt{Correct JSON} lines shown inside it.

\begin{promptblock}[colback=violet!2,colframe=violet!45!black,colbacktitle=violet!45!black]{Demonstration preamble}
Solved examples:
Use these only as demonstrations of the labeling decision. Do not label them again.
\end{promptblock}

\begin{demoblock}{Example 1 --- \texttt{PROVIDE\_REASONING}}
Example 1
Conversation so far:
26. Narwhal: But the king is in check?
27. Narwhal: So they need to move the king or cover with the queen
28. Giraffe: Yeah, wasn't looking there
29. Narwhal: That is about as far as I played it out though
30. Narwhal: I can only go so far ahead haha
31. Hamster: I think black king might be able to make a run for it to d7 for move 3
32. Hamster: but not really sure both moves kinda look winning to me
33. Dolphin: Yeah I'm happy to go with the consenus
34. Narwhal: Ok, happy to compromise as you 3 went for 4
35. Narwhal: Happy to go with 4
Correct JSON:
{"action_label": "PROVIDE_REASONING"}
\end{demoblock}

\begin{demoblock}{Example 2 --- \texttt{EXPLORE\_ALTERNATIVES}}
Example 2
Conversation so far:
28. Emu: damn butterfly.. that's good
29. Panda: 5 looks good if we don't need to play 2, which with check etc we dont need to
30. Emu: i chose 5
31. Alpaca: I choose 5
32. Panda: do we all agree on 5 then?
33. Butterfly: 5.
34. Emu: yep
35. Panda: ok next puzzle
36. Panda: i like 3
37. Emu: chose 2 for this one coz it forces the black to move and gives me an adv
Correct JSON:
{"action_label": "EXPLORE_ALTERNATIVES"}
\end{demoblock}

\begin{demoblock}{Example 3 --- \texttt{REQUEST\_REASONING}}
Example 3
Conversation so far:
130. Beaver: my only thought there is rook to h4
131. Beaver: at the very end
132. Beaver: but i see what you mean
133. Beaver: i would see it as a valid trade off for exposing his king and opening the rook
134. Unicorn: The bishop pins the rook to the king so it can't move
135. Beaver: OH
136. Beaver: my bad
137. Beaver: missed that
138. Beaver: in that case im happy to not do f4
139. Unicorn: yeah, haha
Correct JSON:
{"action_label": "REQUEST_REASONING"}
\end{demoblock}

\begin{demoblock}{Example 4 --- \texttt{EVALUATE\_CRITIQUE}}
Example 4
Conversation so far:
33. Giraffe: Done.
34. Dolphin: Done
35. Butterfly: I think 2 to bring up the pawn tu support the pwan structure?
36. Butterfly: *3 that is
37. Giraffe: Round 3 was more of an arbitrary move for me. I couldn't choose between 1 and 3 so think that I went with 1.
38. Giraffe: Also just wanted to move pawns around
39. Butterfly: Move 1 is fine too but move 3 supports pawn on f and frees up queen
40. Dolphin: I chose 2 to develop the knight more
41. Giraffe: 2 puts the knight in check
42. Butterfly: move 2 gets knight taken by Qf4
Correct JSON:
{"action_label": "EVALUATE_CRITIQUE"}
\end{demoblock}

\begin{demoblock}{Example 5 --- \texttt{COORDINATE\_DECISION}}
Example 5
Conversation so far:
112. Raven: Yeah
113. Raven: I'm 99% sure 3 is winning
114. Leopard: Are you sure we can catch their pawn?
115. Raven: Yep
116. Duck: I think you may be right
117. Leopard: Yes, I have just counted myself. So we would end up with a king, bishop and 2 pawns Vs a king and 2 pawns so we would surely win.
118. Raven: If they push their pawn we just run it down
119. Leopard: Option 3 then?
120. Raven: Then we can eventually promote our pawn to a queen
121. Raven: Yep 3
Correct JSON:
{"action_label": "COORDINATE_DECISION"}
\end{demoblock}

\begin{demoblock}{Example 6 --- \texttt{SOCIAL\_MODERATION}}
Example 6
Conversation so far:
1. Duck: Hello
2. Butterfly: hello
3. Raven: hi
Correct JSON:
{"action_label": "SOCIAL_MODERATION"}
\end{demoblock}

\begin{promptblock}[colback=green!2,colframe=green!40!black,colbacktitle=green!40!black]{Six-shot target block}
Target case:
Conversation so far:
<TARGET DIALOGUE PREFIX>

Output exactly one compact JSON object. Do not include markdown, prose, or explanations.
Return valid JSON only:
{"action_label": "<one allowed label>"}
\end{promptblock}